\documentclass[a4paper,twoside]{article}

\usepackage{epsfig}
\usepackage{subfigure}
\usepackage{calc}
\usepackage{amssymb}
\usepackage{amstext}
\usepackage{amsmath}
\usepackage{amsthm}
\usepackage{multicol}
\usepackage{pslatex}
\usepackage{apalike}
\usepackage{SCITEPRESS}    

\begin{document}

\title{Iris Recognition Under Biologically Troublesome Conditions -\\ Effects of Aging, Diseases and Post-mortem Changes}

\author{\authorname{Mateusz Trokielewicz\sup{1,2}, Adam Czajka\sup{1,3,2} and Piotr Maciejewicz\sup{4}}
\affiliation{\sup{1}Biometrics Laboratory, Research and Academic Computer Network (NASK)\\ Kolska 12, 01-045 Warsaw, Poland}
\affiliation{\sup{2}Institute of Control and Computation Engineering, Warsaw University of Technology\\ Nowowiejska 15/19, 00-665 Warsaw, Poland}
\affiliation{\sup{3}Department of Computer Science and Engineering, University of Notre Dame, IN, USA}
\affiliation{\sup{4}Department of Ophthalmology, Medical University of Warsaw\\ Lindleya 4, 02-005 Warsaw, Poland}
\vskip1mm
\email{(mateusz.trokielewicz, adam.czajka)@nask.pl, piotr.maciejewicz@wum.edu.pl}
}


\abstract{This paper presents the most comprehensive analysis of iris recognition reliability in the occurrence of various biological processes happening naturally and pathologically in the human body, including aging, illnesses, and post-mortem changes to date. Insightful conclusions are offered in relation to all three of these aspects. Extensive regression analysis of the template aging phenomenon shows that differences in pupil dilation, combined with certain quality factors of the sample image and the progression of time itself can significantly degrade recognition accuracy. Impactful effects can also be observed when iris recognition is employed with eyes affected by certain eye pathologies or (even more) with eyes of the deceased subjects. Notably, appropriate databases are delivered to the biometric community to stimulate further research in these utterly important areas of iris biometrics studies. Finally, some open questions are stated to inspire further discussions and research on these important topics. To Authors' best knowledge, this is the only scientific study of iris recognition reliability of such a broad scope and novelty.\let\thefootnote\relax\footnote{Manuscript accepted for publication at the BIOSIGNALS 2017 conference}}

\onecolumn \maketitle \normalsize \vfill

\section{\uppercase{Introduction}}
\label{sec:Intro}
\noindent
Well established position of iris recognition, including several large-scale applications, such as India's Government program AADHAAR, or the CANPASS system maintained for efficient US-Canada border crossings, is attributed to a high uniqueness of the intricate pattern found in the iris tissue, as well as its asserted temporal stability and immutability. This assertion dates back to year 1987 with Safir and Flom's patent, which first laid out theoretical ground for iris recognition: \emph{'significant features of the iris remain extremely stable and do not change over a period of many years'} \cite{SafirFlom}. This is later supported by John Daugman in his 1994 patent, in which he describes the iris pattern as \emph{'unique for each individual and stable over many years'} and \emph{'essentially immutable over a person's life'} \cite{DaugmanPatent}. These claims, being cited throughout the iris biometrics literature, allowed a common belief to arise, that a single enrollment could be sufficient for a lifelong successful recognition of one's identity. 

However, one may come up with several scenarios and circumstances, in which actual iris biometrics performance may tumble short of these perfect-condition assumptions. Recognition accuracy can be heavily influenced by factors related to biological mechanism of the human body. These include \textbf{natural aging} as time progresses, occurence of \textbf{medical conditions and disorders}, and, ultimately, \textbf{death}. 

This paper is intended to summarize Authors' research activity related to the reliability of iris recognition and its resilience against such conditions, as well as to pose some questions regarding these effects' negative impact on recognition accuracy. Sections \ref{sec:Aging}, \ref{sec:Disease} and \ref{sec:ColdIris} present excerpts of comprehensive analyses of effects inflicted by aging, medical disorders affecting the eye, and a \emph{post-mortem} iris recognition study, respectively (with references provided to author's full papers devoted to the respective fields of research). Section \ref{sec:Future} contains conclusions drawn from this study and states some open questions regarding iris recognition performance even under biologically burdensome circumstances, potential gains, but also downsides and risks. 

\section{\uppercase{Iris template non-stationarity}}
\label{sec:Aging}

\paragraph{What is 'template aging'?}
The subject of 'template aging', or, as we think it should be referred to, 'template non-stationarity' in relation to iris recognition can be defined roughly as an increase in error rates that is expected to appear when the time between gallery (\emph{i.e.}, enrollment) and probe (\emph{i.e.}, verification/identification) samples progresses. It consists of multiple aspects that one need to consider for a comprehensive and insightful analysis, including: \emph{biological aging} of the eye and its structures; \emph{differences in sample presentation} originating in pupil dilation, eyelid droop, acquisition conditions, etc.; \emph{sensor interoperability and aging} - when gallery and probe samples are collected using different equipment and camera components wearing out, respectively. The ISO/IEC biometrics vocabulary defines this as follows: \emph{\textbf{reference aging} - the changes in error rates with respect to a fixed reference caused by time-related changes in the biometric characteristic, its presentation, the sensor and other components of the biometric technology}. 

Template non-stationarity is reported to play a vital role in decreasing over-the-years iris recognition performance in number of publications \cite{Gonzalez2008}, \cite{kevin2009factors}, \cite{Baker2009}, \cite{Baker2013}, \cite{Fenker2012}, \cite{Fenker2011}, \cite{FairhurstAging2011}, \cite{Sazonova}, \cite{BowyerOrtiz2015}, including our own previous research \cite{Czajka2013}. NIST's IREX VI report, however, states the contrary \cite{IREX6}, and was later criticized by Bowyer \cite{BowyerIREX}, and a response to that critique was also published \cite{Grother2015}. Recently, more researchers have made efforts to better understand the non-stationarity of templates, namely by isolating as many factors as possible \cite{HofbauerAging2016}, studying the impact of segmentation quality \cite{WildAging2015}, but also the influence of sensor aging \cite{AgingSensorIJCB14}. This shows that despite many research efforts having been put into solving these issues, template aging still presents many challenges, and new solutions and experimental methodologies are much welcome.  

\paragraph{Linear regression analysis.}
We chose to perform an analysis that circumvents the underlying reasons of these changes, and focuses solely on the outcome of the underlying phenomena, manifesting in altering the sample properties. One way to do so is to perform a linear regression analysis that aims at predicting the comparison score using covariates relating to some predefined qualities of the image. This is to show possible sources of the decrease in recognition accuracy as time progresses \cite{TrokielewiczAgingIWBF2015}. 

For the experiments, a database of 583 samples from 58 irises collected up to 9 years apart have been used (to our knowledge this is one of the most extensive aging-related iris images databases in terms of the timespan between samples). Linear regression analysis was employed in attempt to predict the comparison score in terms of: \textbf{1) time elapsed} since gallery image acquisition, \textbf{2) selected quality measures} (eyelid, eyelash and reflection occlusion percentage, local contrast, illumination intensity and image sharpness) and \textbf{3) geometrical factors} (iris and pupil radii and their variability in a given image pair). 29 regression models built upon three different commercial and academic iris recognition solutions allowed us to formulate some interesting conclusions.

The time parameter proved to be statistically significant in every model, making it plausible that the non-stationarity phenomenon may be autonomous from quality and geometrical characteristics of iris images. Nonetheless, those covariates should be taken into account in future studies, as some combinations of them turn out to also be statistically significant in predicting the comparison score, such as image sharpness and local contrast. Notably, the pupil-to-iris radius ratios are shown to be statistically significant in every tested model. This may indicate that differences in pupil diameter are the most likely sources of recognition accuracy decrease as time between sample acquisitions elapses.

\section{\uppercase{Iris biometrics and ocular disorders}}
\label{sec:Disease}

\paragraph{Recognition accuracy impact.}
Iris recognition usually performs exceptionally well, provided that it is applied to subjects with healthy eyes. However, numerous medical conditions affecting the eye structures, especially the iris, anterior chamber of the eye, and the cornea, have a potential of degrading its accuracy and reliability. Yet, due to the lack of appropriate datasets and difficulties in creating them, limited research is available, mostly centered around cataract and cataract extraction procedure influence on iris recognition performance: \cite{Roizenblatt}, \cite{Seyeddain2014}, \cite{Dhir}, \cite{TrokielewiczWilga2014}, \cite{RamachandraCataractICB2016} (significant negative impact of cataract and cataract surgery reported by most researchers except for Dhir \emph{et al.}), impact of refraction correction procedures \cite{Yuan} (no impact reported), but also studies regarding multiple disorders \cite{Aslam}, and their impact on segmentation \cite{McConnon2012}. In the papers \cite{TrokielewiczCYBCONF2015}, \cite{TrokielewiczBTAS2015} we present the most thorough and comprehensive analysis on the subject of disease influence on iris recognition reliability to date, including an extensive cataract influence study, and a novel approach to eye pathology impact analysis, based not on disease taxonomy (impact of certain diseases), but rather on the type of damage that medical disorders afflict on the eye. 

\paragraph{Database of iris images collected from ophthalmology patients.}
For the purpose of these studies, a new database had to be collected. We had a rare opportunity of a close collaboration with an ophthalmologist's office, which allowed us to gather an unprecedented collection of iris images coming from patients suffering from more than 20 different conditions. This dataset consists of almost 3000 images, both NIR-illuminated and high-resolution ones taken in visible light (this is done to enable a close-up visual inspection of the affected eye structures). 

\paragraph{Cataract.} Our research devoted to studying the influence of cataract on the performance of various iris recognition methods showed a degradation in matchers' accuracy when images obtained from cataract affected eyes are used, compared to the scenario when images of healthy eyes are used. This decrease in performance manifested itself with worsening the genuine comparison scores by as much as 175\% (for an example commercial matcher), while impostor scores remained mostly unaffected. This change in comparison scores was able to elevate the FNMR values in two out of three employed recognition methods. Additional experiments were conducted to show whether this decrease in performance could be attributed with wrong execution of the image segmentation stage, however, this hypothesis was not confirmed. Hence, we may suspect that some additional factors play a vital role in worsening the reliability of iris recognition in cataract patients \cite{TrokielewiczWilga2014}.   

\paragraph{Disease impact on eye structures.} While this is often true for cataract-affected eyes, most ophthalmology patients with severe eye illnesses suffer from not one, but usually two or more conditions at the same time. These conditions are often unrelated and affecting the eye in different ways. This makes it extremely difficult to conduct an insightful analysis for one disease at a time. Hence, we proposed a new method of data analysis, which involves dividing the dataset into five subsets, each of them representing a different type of impact afflicted on the eye structures by the pathologies involved. The five partitions include: 1) healthy eyes, 2) disease-affected eyes, but not revealing any visible impairments, 3) eyes with geometrical distortions of the pupil, 4) eyes with changes in the iris tissue itself, and 5) eyes with changes in the cornea or the anterior chamber that obstruct the view of the iris below those structures. This approach, combined with an exhaustive analysis incorporating four commercial and academic matchers, allowed us to formulate four interesting and important conclusions \cite{TrokielewiczCYBCONF2015}, \cite{TrokielewiczBTAS2015}, \cite{TrokielewiczDiseasesIMAVIS}:

\begin{itemize}
	\item {\bf the enrollment stage is highly sensitive to medical conditions that introduce geometrical distortions to the pupillary area and obstructions of the iris pattern}
	\item {\bf even if no perceivable changes can be observed in the diseased eyes, the performance can still drop when compared to this achieved using healthy eyes images}
	\item {\bf all eye conditions that can afflict visible damage to the eye structures are capable of degrading the comparison scores (across all tested recognition methods), with geometrical deformations and iris pattern obstructions contributing the most}
	\item {\bf most of the observed recognition errors can be attributed to the faulty execution of the image segmentation stage}
\end{itemize}

\paragraph{Database contribution.} Papers \cite{TrokielewiczCYBCONF2015} and \cite{TrokielewiczBTAS2015} also make a significant contribution for the biometrics community by offering two vast datasets of iris images obtained from patients suffering from various ocular disorders. We are not aware of any other publicly available datasets that would offer a collection of iris images representing disease-affected eyes. Those datasets can be used for research and non-commercial purposes by all interested researchers. For details on how to access the data, please see \cite{WarsawDiseaseIris1}, \cite{WarsawDiseaseIris2}.

\vfill

\section{\uppercase{Post-mortem iris recognition}}
\label{sec:ColdIris}
\paragraph{A benefit for forensics, an issue for identity management?}
The topic of post-mortem recognition in human subjects has received considerably low attention in the biometric community. Due to the difficulties in data collection and the obvious unpleasantness of such experiments, very little research has been published, especially when human eyes are concerned, with few exceptions, namely \cite{BostonPostMortem} (the paper concludes that post-mortem iris recognition works fine in about 80\% of the cases for samples acquired up to 2 days after death) and  \cite{PostMortemBoehnenBTAS16} (which mostly focuses on post-mortem face and fingerprint recognition, with few conclusions regarding irises). \cite{PostMortemPigs} present a study of post-mortem iris recognition using cadaver eyes of a domestic pig, reporting that the eyes lose their capability to serve as a biometric identifier in 6 to 8 hours post-mortem. 

This aspect of iris biometrics is important for at least two reasons. First, if post-mortem recognition is viable, it could prove useful in forensics, namely identification and verification of accident and crime victims, and even in the battlefield (when other fast methods of identification are not accessible, say, victim has lost his fingers or face is disfigured). The latter reason connects with the use of iris biometrics for identity management and asset protection and an associated fear of identity theft - \emph{'will someone be able to steal my iris after I die, and use it to gain access to my identity?'} \cite{ScienceFocusPostMortem}. Several publications firmly mention that iris recognition after death cannot be performed due to pupil dilation and corneal cloudiness \cite{DaugmanPostMortem}, \emph{'iris decay'} \cite{SaeedPostMortem}, \emph{'iris features vanishing with pupil dilation'} and \emph{'muscle relaxation'} \cite{IrisGuardPostMortem}\cite{IriTechPostMortem}. However, no experimental evidence is presented in either of those publications. 

\paragraph{Experimental study: short-term analysis.} 
 In our studies regarding the field of post-mortem iris biometrics \cite{TrokielewiczPostMortemICB2016}, \cite{TrokielewiczPostMortemBTAS2016} we have shown that the above claims are mostly untrue. To be able to conduct these experiments, a new database had to be collected, using iris images obtained from deceased human subjects in a hospital mortuary. The dataset comprises of iris images collected from 12 different irises over a period of 27-29 hours post-mortem. The first session was conducted approximately 5-7 hours after demise, with the second and third sessions conducted after 11 and 22 hours. We managed to show that, contrary to claims cited above, the pupils are not excessively dilated after death (but rather fixed in a mid-dilated position), nor is the iris structure \emph{'vanishing'}. With images captured a few hours post-mortem we were able to reach perfect recognition accuracy with one of the four employed matchers, while the FNMR values for the remaining three were surprisingly low (from 1.4\% to 8.3\%). The decay of the eye structures indeed progresses as time after death elapses, yet these dynamics are much less aggressive than previously stated in literature. The FNMR values for the best performing matcher rose to 5.1\% and 26.7\% for images obtained in the second and the third session, respectively (images from these sessions were compared against those obtained in the first session). It is the third session, with images collected approximately 27 hours post-demise, where serious deterioration begins and error rates spike, depending on the method employed, to the range of 26.7\%-86.7\% of falsely non-matched samples.
 \paragraph{Experimental study: long-term analysis.} 
 Following these studies, we have continued to collect the data and were able to obtain a unique dataset of post-mortem human iris images spanning as long as 407 hours (almost 17 days). These experiments revealed that although after such a long period iris recognition is almost impossible, one may still expect to get occasional correct matches (after 407 hours for IriCore and MIRLIN methods, 260 hours for the VeriEye method, and 124 hours for the OSIRIS solution). However, apart from these exceptions, the rate of eye degradation due to drying, wrinkling, and opacification of the cornea combined with a collapse of the eyeball make iris recognition virtually impossible in most of the attempts after such long periods after death. 

\paragraph{Database contribution.} Notably, here as well a unique dataset of iris images obtained from deceased subjects has been prepared and will be released to interested members of the biometric community in the fall, to encourage further research on this important matter. To author's best knowledge, this will also be the first publicly available dataset of this kind. On how to get access to the data, please see \cite{WarsawColdIris1}.  

\vfill

\section{\uppercase{Conclusions and open questions}}
\label{sec:Future}
This paper presents a comprehensive account on Authors' research regarding iris recognition's still untackled problems associated with biological processes taking place in the human body. Aging, ocular pathologies, and processes occurring after death are shown to be capable of causing serious degradation in the reliability of various commercial and academic iris recognition solutions. Although the very existence of these issues should not by any means lead to dismissing iris biometrics as a secure, efficient and accurate identification method, certain steps should be undertaken to defend against them. Thus, an important aspect of future studies should be to propose appropriate countermeasures that would increase resilience of iris recognition against changes induced in the eye by these phenomena. 

The questions that arise from this research, are thus as following:
\begin{itemize}
	\item what contributes to the \emph{iris template non-stationarity} phenomenon? What methods can be employed to examine these effects?
	\item how can we defend against decrease in recognition accuracy caused by biologically-induced damage to the eye, such as diseases and post-mortem decay?
	\item can \emph{post-mortem} iris recognition provide a new method to improve over currently used toolboxes of forensic examiners?
	\item on the other hand, should \emph{post-mortem} iris recognition pose concerns over biometric identity management security (are there vulnerabilities, such as presentation attack risks)?
\end{itemize}

We hope that this paper, together with the available datasets, will inspire other researchers in this field to come up with their own experiments regarding the interdisciplinary field of biometrics and biology, and solutions to problems discussed here, to further improve iris recognition as a safe, fast, and reliable biometric method.

\bibliographystyle{apalike}
{\small
\bibliography{refs}}

\end{document}